\documentclass[conference]{IEEEtran}
\IEEEoverridecommandlockouts
\usepackage{cite}
\usepackage{amsmath,amssymb,amsfonts}
\usepackage{algorithmic}
\usepackage{graphicx}
\usepackage{textcomp}
\usepackage{xcolor}
\usepackage{multirow}
\usepackage{boldline}
\usepackage{stackengine}
\usepackage{siunitx}
\usepackage{physics}
\usepackage{booktabs}

\def\BibTeX{{\rm B\kern-.05em{\sc i\kern-.025em b}\kern-.08em
    T\kern-.1667em\lower.7ex\hbox{E}\kern-.125emX}}
\begin{document}

\title{Feature-weighted Stacking for Nonseasonal Time Series Forecasts: A Case Study of the COVID-19 Epidemic Curves}

\author{
\IEEEauthorblockN{1\textsuperscript{st} Pieter Cawood}
\IEEEauthorblockA{\textit{Computer Science and Applied Maths} \\
\textit{University of the Witwatersrand}\\
Johannesburg, South Africa \\
pieter.cawood@gmail.com}
\and
\IEEEauthorblockN{2\textsuperscript{nd} Terence L. van Zyl}
\IEEEauthorblockA{\textit{Institute for Intelligent Systems} \\
\textit{University of Johannesburg}\\
Johannesburg, South Africa \\
tvanzyl@gmail.com}
}

\maketitle

\begin{abstract}
We investigate ensembling techniques in forecasting and examine their potential for use in nonseasonal time-series similar to those in the early days of the COVID-19 pandemic. Developing improved forecast methods is essential as they provide data-driven decisions to organisations and decision-makers during critical phases. We propose using late data fusion, using a stacked ensemble of two forecasting models and two meta-features that prove their predictive power during a preliminary forecasting stage. The final ensembles include a Prophet and long short term memory (LSTM) neural network as base models. The base models are combined by a multilayer perceptron (MLP), taking into account meta-features that indicate the highest correlation with each base model's forecast accuracy. We further show that the inclusion of meta-features generally improves the ensemble's forecast accuracy across two forecast horizons of seven and fourteen days. This research reinforces previous work and demonstrates the value of combining traditional statistical models with deep learning models to produce more accurate forecast models for time-series from different domains and seasonality. 
\end{abstract}


%%%%%%%%%%%%%%%%%%%%%%%%%%%%%%%%%%%%%%%%%%%%%%%%%%%%%%%%%%%%%%%
%                        INTRODUCTION
%%%%%%%%%%%%%%%%%%%%%%%%%%%%%%%%%%%%%%%%%%%%%%%%%%%%%%%%%%%%%%%

\section{Introduction}
With a socially integrated planet, breakouts of pandemics like SARS, Ebola and the recent COVID-19 have catastrophic effects on the wellness and economies of countries~\cite{van2021did}. During a pandemic, case numbers are recorded cumulatively for a specific country or region. This recording results in a univariate time-series with a daily frequency. Governments and organisations utilise this information for simulation and forecasting techniques to inform decision making during pandemics.

Despite the immense diversity of the forecasting models~\cite{perumal2020comparison}, the no free lunch theorem \cite{wolpert1997no} holds for forecasting that no single algorithm universally outperforms all others for all problems. For this reason, ensembling is an attractive state of the art approach that combines the forecasts from numerous models to obtain improved predictive performance. 

For instance, a model averaging ensemble is currently used for COVID-19 forecasting by the United States Centers for Disease Control and Prevention (CDC) \footnote{https://www.cdc.gov/coronavirus/2019-ncov/science/forecasting/forecasting-us.html}. However, a stacked generalisation approach such as the averaging ensemble is not guaranteed to improve forecast accuracy as noted by Ribeiro \emph{et al.} (2020) \cite{ribeiro2020short}. The work of Ribeiro \emph{et al.} (2020) \cite{ribeiro2020short} is likely the most closely related, who proposed to forecast cumulative Brazilian COVID-19 case numbers using four machine learning methods and to have a Gaussian process (GP) combine their results to produce the final forecasts.

As the demand for forecasts in business domains~\cite{atherfold2020method} is higher than that of epidemiology, research delivering improved forecasting techniques are standard for financial- and seasonal time-series \cite{MakridakisComparisons, ribeiro2020ensemble, makridakis2020m4}. However, there is relatively less research dedicated to ensembles of non-seasonal epidemic curves. To this end we include meta-features in a stacking design as proposed by Coscratoa \emph{et al.} (2020) \cite{coscrato2020nn}. 

By learning from meta-features, the meta-learner has the advantage of learning how to weight distinctive regions of the feature space concerning how reliable the individual forecast models are. The feature-weighted component relates to research that improves an ensemble's accuracy by including meta-features in the design. Most closely related is the feature-based forecast model averaging (FFORMA) framework proposed by Montero-Manso \emph{et al.} (2019) \cite{montero2020fforma}. Their framework allows a meta-learning model to produce weights for a pool of methods based on their learned performance observed for some time-series statistics (meta-features). Our research considers the case of the novel Coronavirus (COVID-19) epidemic curves to evaluate the models for multiple countries with differently characterised curves.

We contribute to the forecasting research by:
\begin{itemize}
    \item demonstrating that characteristics of a nonseasonal time-series can be learned and, as a result, improve the predictive power of neural network ensemble models,
    \item evaluating time-series statistics as meta-features and show their utility in improving forecasting with ensembles, and
    \item exploring the potential for different forecasting ensemble techniques within the nonseasonal epidemic domain.
\end{itemize}

\section{Data, Models and Methodology}

%%%%%%%%%%%%%%%%%%%%%%%%%%%%%%%%%%%%%%%%%%%%%%%%%%%%%%%%%%%%%%%
%                            DATA 
%%%%%%%%%%%%%%%%%%%%%%%%%%%%%%%%%%%%%%%%%%%%%%%%%%%%%%%%%%%%%%%

\subsection{Data}

\subsubsection{Description}
We used a single source as a global COVID-19 data set, which is shared freely by researchers from the John Hopkins University \cite{dong2020interactive}. The data set includes $215$ countries' case numbers from mid-January 2020 and is updated daily on a GitHub repository \footnote{https://github.com/CSSEGISandData/COVID-19}. The forecasts in this paper utilise the cumulative case numbers of the" Confirmed" counts of $21$ countries as the independent variables.  The "Date" feature is chosen as the dependent variable, allowing us to use the data as a nonseasonal- univariate time-series. Note that this is a relatively small set of data, and therefore, more work is needed to evaluate the superiority of the proposed method for more extensive data sets.

\subsubsection{Pre-processing and decisions}\label{sec::split}
\begin{figure}[!ht]
	\centering
	\includegraphics[width=\columnwidth]{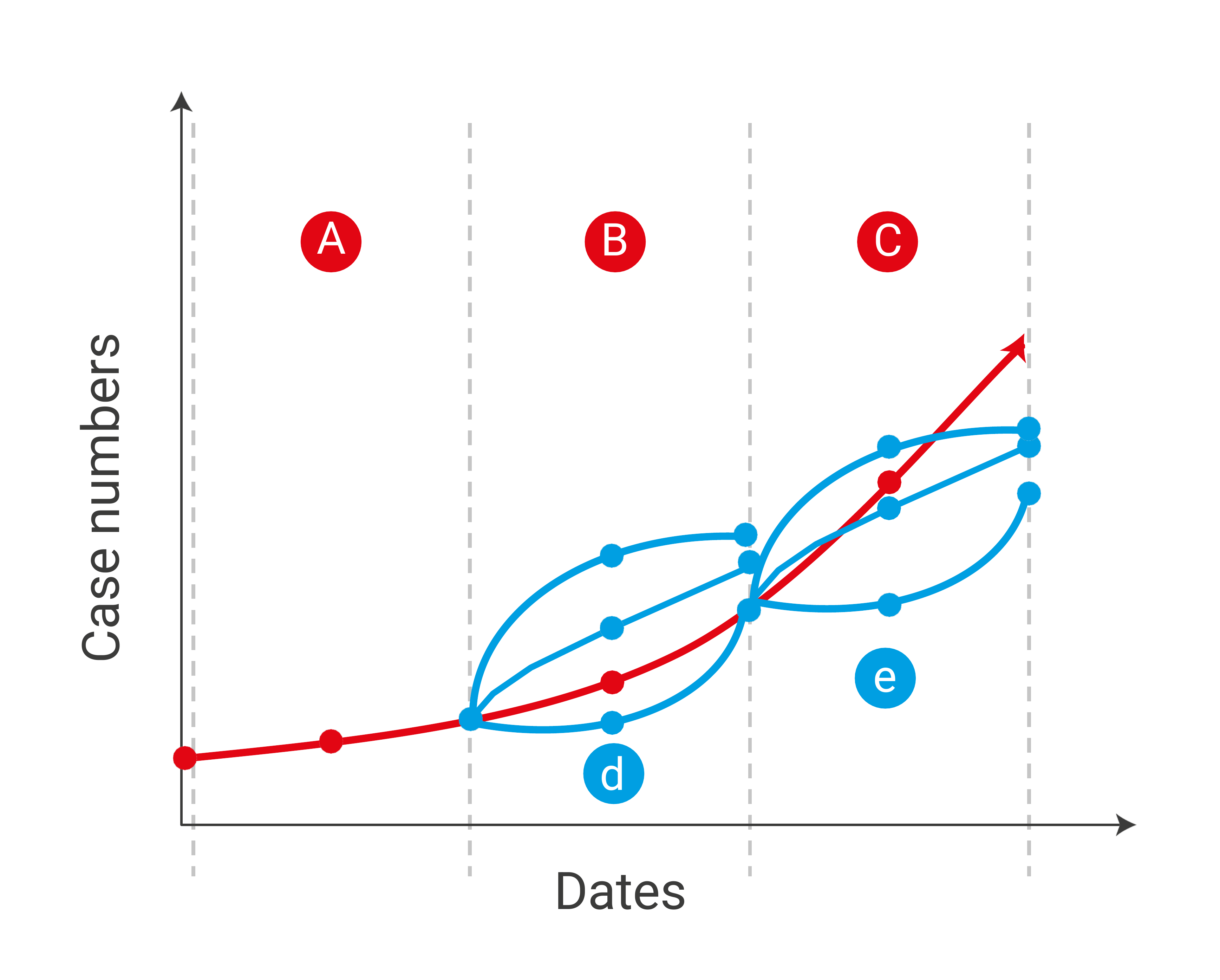}
\resizebox{\columnwidth}{!}{
{
\begin{tabular}{ll}
&\\
A                  & First window of the curve used for the training of the base models.\\[4pt]
\multirow{2}{*}{B} & Second window of the curve used as targets for forecasts with window A   \\
                   & and an additional set of training curves.                           \\[4pt]
C                  & Final window of the curve used as targets for window B. \\[4pt]
d                  & Base model outputs used for the training of the meta-learners. \\[4pt]
\multirow{2}{*}{e} & Base model outputs used for the training of the meta-learners.  
\end{tabular}
}}
\caption{Data splitting used for each time-series.}\label{fig_data_split}
\end{figure}

We use a Box-Cox transformation to improve the time-series stationarity and remove some of the random variation in the time-series signal. Additional smoothing was performed using a moving average with a window size of two days. We do model parameterisation using multiple $30$ days of historical observations. The models are evaluated for unseen data of one short term forecast of seven days and a longer horizon forecast of fourteen days,  which is represented by the "C"-labelled section of figure \ref{fig_data_split}. 

Twenty-nine countries were chosen from the data set; that deliver $261$ subsets of curves after applying the proposed time-splitting procedure. The countries include Australia, Algeria, Brazil, France, Germany, India, Italy, Japan, Kenya, Mexico, Poland, Russia, South Africa, Turkey, USA, Peru, Lebanon, Chile, Bangladesh, France and the UK for the training curves and Saudi Arabia, Canada, Portugal, Egypt, Belgium, Netherlands, Canada and Sweden for out-of-sample forecasts. The first batch of $21$ countries is used during the training stage to make the "d" and "e" forecasts of figure \ref{fig_data_split}. The meta-learners make their final predictions using the "e" section of the second batch of altogether eight held out countries.

\subsection{Meta-features}
Statistics that describe the time-series' complexity and behaviour were considered after reviewing the proposed features for a stacking ensemble to forecast the data of the M3 Makridakis forecasting competition \cite{barak2019time}. 

The coefficient of variation ($\operatorname{CV}$) was selected as a measurement of the time-series' smoothness, using a ratio found by:
\begin{equation}
    \operatorname{CV} = \frac{\sigma}{\mu}
\end{equation}
where $\mu$ is the mean value and $\sigma$ is the standard deviation.

Singular value decomposition entropy (SVD entropy) was chosen as a statistical measurement of entropy \cite{caraiani2014predictive} to determine whether measurement of the dimensionality might help predict each model's accuracy. The SVD entropy is defined as:
\begin{equation}
    H = - \sum{ \lambda_k \ln (\lambda_k) } 
\end{equation}
where $\lambda_k$ denotes the singular values which might be found for a matrix of produced from the observed values.

\par The Kwiatkowski–Phillips–Schmidt–Shin (KPSS) test \cite{kwiatkowski1992testing} provides a measure of the time series' stationarity around the linear trend. The KPSS test is based on linear regression and it is found with three sub-components as define by
\begin{equation}
    \operatorname{x_t} = r_t + \beta_t + \epsilon_t
\end{equation}
where $r_t$ is the deterministic trend, $\beta_t$ is the random walk and $\epsilon_t$ is the stationary error.

The autocorrelation function (ACF) measures the level of non-randomness in the time series by computing the correlation of the time series with a lagged copy of itself. The $\operatorname{ACF}$ is defined as
\begin{equation}
    \operatorname{r_k} = \sum_{i=1}^{N-k}\frac{(Y_i - \bar{Y})(Y_{i+k} - \bar{Y})}
                                              { \sum_{i=1}^N(Y_i - \bar{Y})^2}
\end{equation}
where $k$ denotes the length of the lag, $X$ is the independent time variable, $Y$ is the observed value and $\bar{Y}$ is the observed value of the lagged series.

\subsection{Metrics}
We use Spearman's rank correlation coefficient $\rho$ as a metric, which allows us to measure the relationship between the meta-features values and the performance of each base model. This metric allows us to analyse the data to select the two meta-features that have the highest correlation with the accuracy scores of the two best performing base models.

In addition, the symmetric mean absolute percentage error ($\operatorname{sMAPE}$) is used as the accuracy measurement during optimisation and testing stages, and it is denoted by:
\begin{equation}
\operatorname{sMAPE} = \frac{1}{n}\sum_{t=1}^{n}{2\frac{\abs{F_t+A_t}}{(\abs{A_t}+\abs{F_t})}}
\end{equation}
where $F_t$ is the forecast value and $A_t$ is the actual value.

%%%%%%%%%%%%%%%%%%%%%%%%%%%%%%%%%%%%%%%%%%%%%%%%%%%%%%%%%%%%%%%
%                        BASE MODELS 
%%%%%%%%%%%%%%%%%%%%%%%%%%%%%%%%%%%%%%%%%%%%%%%%%%%%%%%%%%%%%%%

\subsection{Forecasting models}
We use four different forecast methods as the potential base models for a machine learning ensemble. A brief explanation of each forecasting method follows.

\subsubsection{ARIMA}
The autoregressive integrated moving average (ARIMA) model is a classical- and perhaps still one of the most prominently used additive methods for time-series forecasting. The method uses three iterative sub-processes to perform the forecasting of nonstationary time-series. Fattah \emph{et al.} (2018) \cite{fattah2018forecasting} describes the ARIMA model as a Box-Jenkins approach defined by:
\begin{align}
    \operatorname{ARIMA}(p,d,q)
\end{align}
where $p$ is a number of autoregressive terms, $d$ is the number of differences and $q$ is a parameter of the number of moving average, and the processes are repeated until the parameters that cause the lowest errors are determined.

\subsubsection{HW}
The Holt-Winters (HW) is another classical approach to forecasting using exponential smoothing. We use double exponential smoothing \cite{kalekar2004time} that acts as an additive model. The HW adds two exponentially decreasing weights to perform forecasting implemented using the following two equations:
\begin{align}
S_t = \alpha * y_t + (1- \alpha) * (S_{t-1} + b_{t-1}) \;\;\; 0 < \alpha < 1 \\
b_t = \gamma * (S_t - S_{t-1}) + (1 - \gamma) * b_{t-1} \;\;\; 0 < \gamma < 1
\end{align}
where $\alpha$ and $\gamma$ are constants to each smoothing process.

\subsubsection{Prophet}
Prophet is a collection of models developed by researchers at Facebook \cite{taylor2018forecasting}. Prophet is also an additive model that fits a function of time as the trend, seasonality and holidays components. The linear trend with the changepoints method was implemented, allowing linear trends using a piece-wise constant rate of growth. The method is obtained with:
\begin{align}
g(t) = (k + a(t)^T\delta)t + (m + a(t)^T)
\end{align}
where $k$ denotes the growth rate, $\delta$ is the rate of adjustments, $m$ an offset parameter and $\gamma$ is used to make the function continuous.

\subsubsection{LSTM}
The final model is a deep learning forecasting model~\cite{mathonsi2020prediction}. The long short-term memory (LSTM) extends the recurrent neural network (RNN) architecture by implementing gated memory blocks in the recurrent hidden layer. LSTM was originally introduced to mitigate the vanishing gradient problem in the classical RNN. Sak \emph{et al.} (2020) \cite{sak2014long} described the LSTM networks’ computation as a mapping from the input sequence $x = (x_1, \dots , x_T ) $ to the output sequence $(y_1, \dots , y_t )$, and the network activations might be found by iterating the following equations from $t = 1$ to $T$:
 \begin{equation}
 \begin{aligned}
i_t = \sigma(W_{ix}x_t + W_{im}m_{t-1} + W_{ic}c{t-1} + b_i)  \\
f_t = \sigma(W_{fx}x_t + W_{fm}m_{t-1} + W_{fc}c{t-1} + b_f) \\
c_t =  f_t 	\odot c_{t-1} + i_t \odot g(W_{cx}x_t + W_{cm}m_{t-1} + b_c) \\
o_t = \sigma (W_{ox}x_t + W_{om}m_{t-1} + W_{oc} c_t + b_o) \\
m_t = o_t \odot h(c_t) \\
y_t =  \phi (W_{ym}m_t + b_y)
\end{aligned}
\end{equation}
where $\sigma$ is the logistic sigmoid function. $W_{ix}$, $W_{ox}$ and $W_{cx}$ terms are weight matrices from the input gates to the inputs. $W_{im}$, $W_{om}$ and $W_{cm}$ are diagonal weight matrices as peephole connections. $m$ is the output vector of the memory block, and the $b$ terms denote a bias vector on each gate type. $g$ and $h$ are the cell input and output activation functions, respectively, and $\phi$ is the activation function used for the network. $c_t$ is the cell state vector, and $\odot$ denotes the element-wise product of vectors.

%%%%%%%%%%%%%%%%%%%%%%%%%%%%%%%%%%%%%%%%%%%%%%%%%%%%%%%%%%%%%%%
%                        META-LEARNER
%%%%%%%%%%%%%%%%%%%%%%%%%%%%%%%%%%%%%%%%%%%%%%%%%%%%%%%%%%%%%%%
\begin{figure*}[!hbt]
  \includegraphics[width=\textwidth,height=12cm]{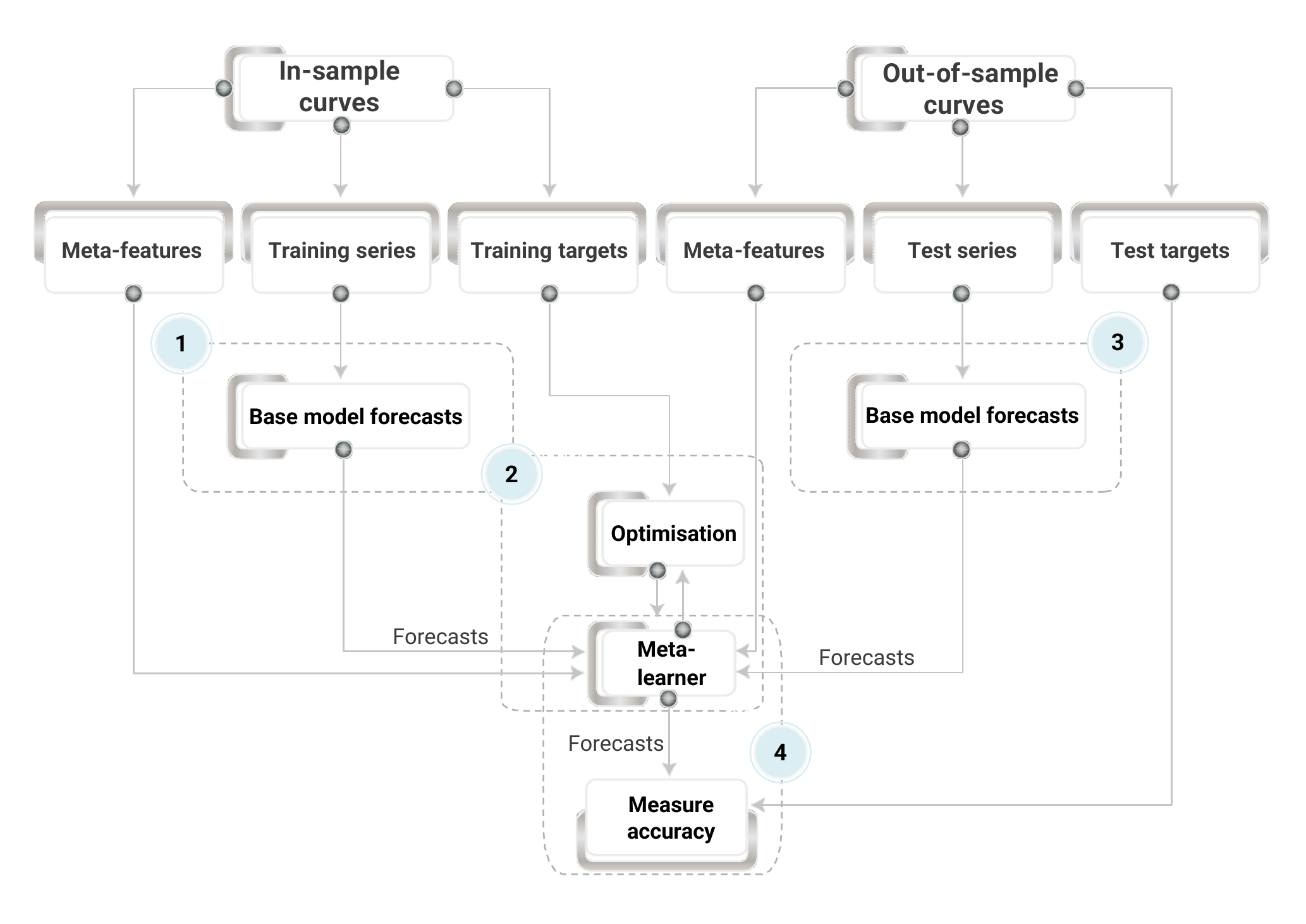}
  \caption{The proposed feature-weighted stacking architecture.}
  \label{fig::architecture}
\end{figure*}

\subsection{Meta-regression}
More traditional stacking ensembles learn only by performing regression on the predictions of the base models. In the meta-regression case, we use the outputs from the selected base models together with meta-features to allow the fitting and predictions using any regression algorithm. This paper describes using a multi-layer perceptron (MLP) as the meta-learner, as neural networks are capable of nonlinear modelling and they are known to be good universal approximators. 

For the first batch of $21$ countries' curves, the forecasts from the base models are supplied with their meta-features as inputs to train the MLP meta-learner. Only the second section of the eight held out countries' curves are used (with their meta-features) to make the final forecasts. The MLP model is generally composed using at least three layers of neurons which allows for a function that maps the inputs to a non-linear output \cite{agirre2006regression} using the following equation:
\begin{align}
y^o_k &= f^o_k(b^o_k + \sum_{i=1}^{S} w^o_{ik} f^h_i( b^h_i + \sum_{j=1}^{N} w^h_{ji}x_j))
\end{align}
where $N$ denotes the number of neurons in the input layer, $S$ denotes the number of neurons in the hidden layer, $h$ denotes the units of the hidden layer, $o$ is the number of units of the output layer, $b$ is the bias of the neuron and $f$ denotes the network activation function.

By following this iterative equation, the MLP model can learn and weight the base models according to their learned results for varying characteristics of the time-series and make final weighted forecasts using a single output layer. 

The inputs to the meta-learner are a vector concatenation of the ensembles' predictions at time-step $t$ and the meta-features extracted for the reference series. This concatenation allows the meta-learner to learn from the learning process of each base model as observed for differently characterised time-series (meta-feature measurements).

\subsection{Proposed forecasting architecture}
Figure \ref{fig::architecture} depicts the proposed feature-weighted stacking architecture, which is achieved with the following main steps:
 \begin{description}
	\item[\textbf{Step 1:} ] \ The data is split using the proposed time splitting technique to create two sets of curves for each in-sample country. All of the base models then make preliminary forecasts for each curve. The forecast values and meta-features that describe their input curves are supplied to the next step.
    \item[\textbf{Step 2:} ] \ The Spearman's rank correlation coefficient and sMAPE scores are computed for all of the preliminary forecasts, and the two best performing models are used as the base models in the ensemble, with the two highest correlating meta-features. The selected model forecasts and meta-features are then used to firstly optimise the meta-learner's hyper-parameters with the actual target values for each forecast. The same set of forecasts, meta-features and target values are then used to train the optimised Meta-learner.
    \item[\textbf{Step 3:} ] \  The selected base models make forecasts from the out-of-sample countries, and their forecast values and meta-features that describe their input curves are supplied to the next step.
    \item[\textbf{Step 4:} ] \ The meta-learner makes the final forecasts using the prediction values and meta-features from step 3 as inputs and the ensemble's accuracy is measured.
 \end{description}

%%%%%%%%%%%%%%%%%%%%%%%%%%%%%%%%%%%%%%%%%%%%%%%%%%%%%%%%%%%%%%%%
%                         RESULTS
%%%%%%%%%%%%%%%%%%%%%%%%%%%%%%%%%%%%%%%%%%%%%%%%%%%%%%%%%%%%%%%
	
\section{Experimental results}

\begin{figure*}[!hbt]
    \centering
    \includegraphics[width=0.75\textwidth]{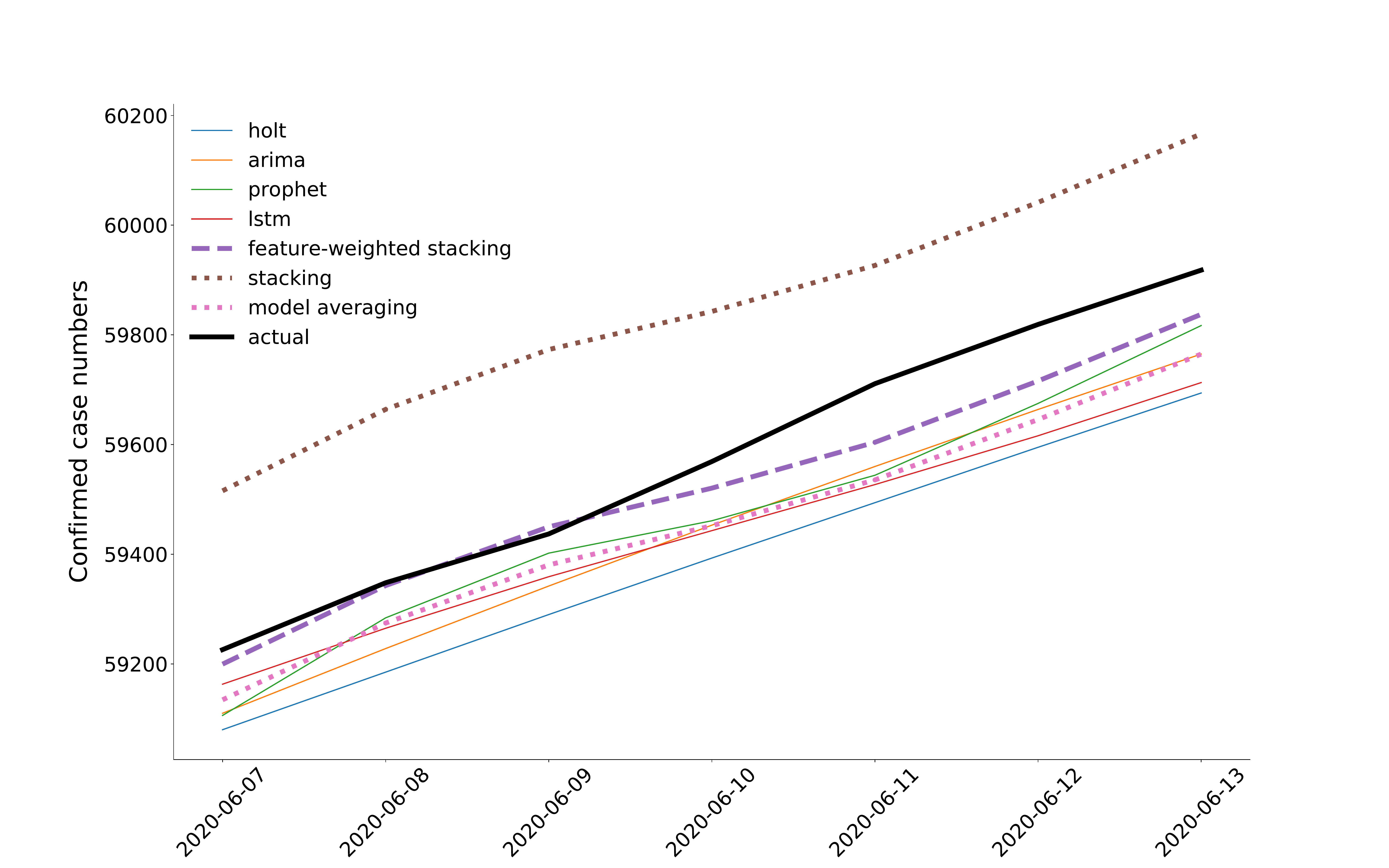}
    \caption{Seven-day forecasts for Egypt.}
    \label{fig::egypt}
\end{figure*}

\begin{figure*}[!hbt]
    \centering
    \includegraphics[width=0.75\textwidth]{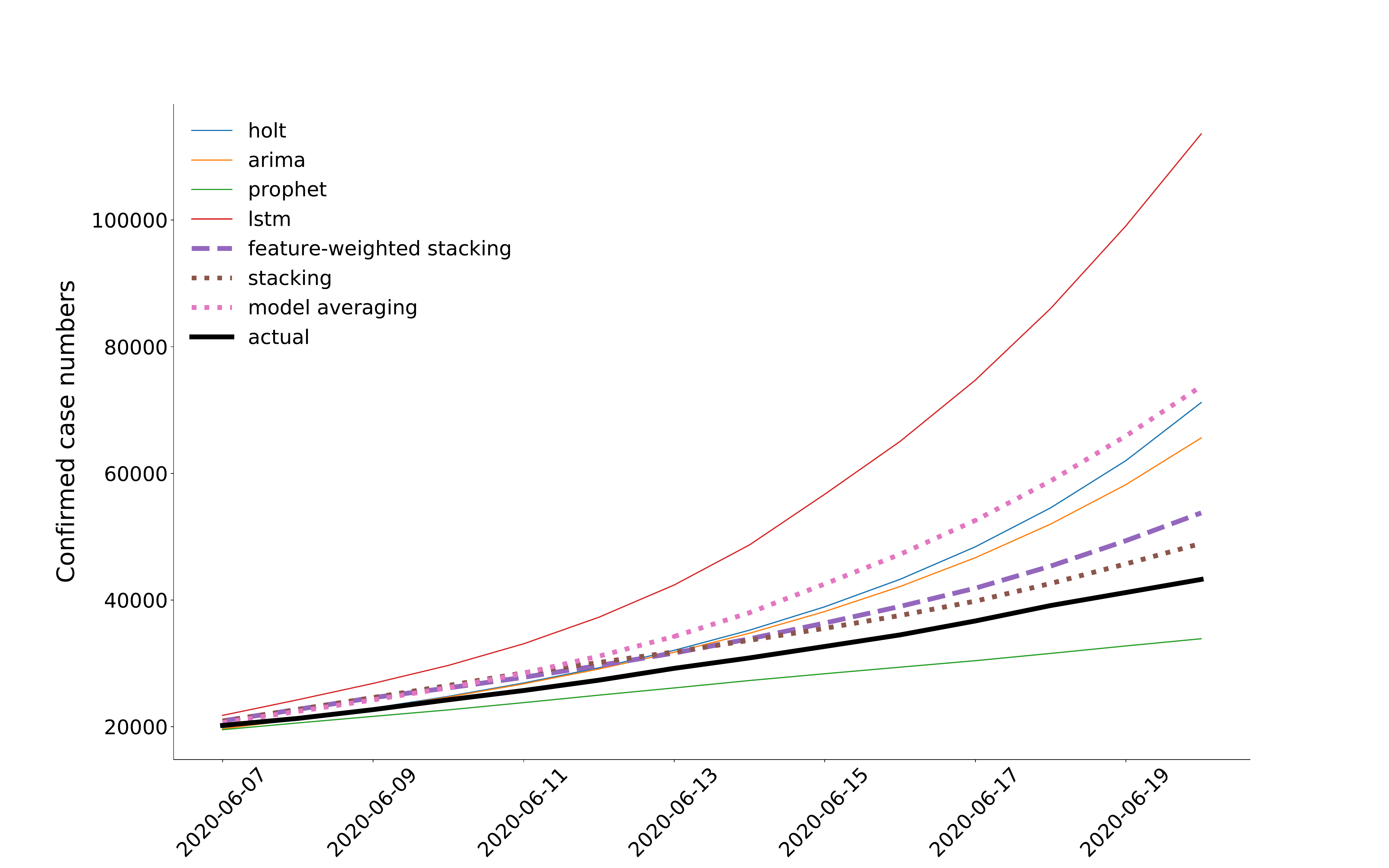}
    \caption{Fourteen-day forecasts for the Netherlands.}
    \label{fig::netherlands}
\end{figure*}

\subsection{Meta-feature correlation}
The meta-features was extracted from all curves used in the first batch of curves. The average Spearman's rank correlation coefficient ($\rho$) was used to measure their correlation with the achieved sMAPE scores. The results for both forecasting horizons are captured in Table~\ref{tab:table_spearman_7} and Table~\ref{tab:table_spearman_14} respectively. This novel approach allows us to choose meta-features for our ensemble that show the best potential for forecasting the expected outcomes.

\begin{table}[!ht]
\centering
\caption{Spearman's rank correlation coefficient found for each base model's 7 days forecasts.}
\resizebox{0.8\columnwidth}{!}{
{
\begin{tabular}{l|c|c|c|c}
                              & CV & SVDE & KPSS & ACF   \\ 
\bottomrule\toprule
\multicolumn{1}{l|}{ARIMA}   & 0.46 & \multicolumn{1}{c|}{0.25}  & 0.4 & \multicolumn{1}{c}{0.24} \\
\multicolumn{1}{l|}{HW}      & 0.46 & \multicolumn{1}{c|}{0.29} & 0.22 & \multicolumn{1}{c}{0.01} \\
\multicolumn{1}{l|}{Prophet} & 0.45 & \multicolumn{1}{c|}{0.23} & 0.31 & \multicolumn{1}{c}{0.12} \\
\multicolumn{1}{l|}{LSTM}    & 0.45 & \multicolumn{1}{c|}{0.23} & 0.42 & \multicolumn{1}{c}{0.13} \\ 
\bottomrule
\end{tabular}
}
}
\label{tab:table_spearman_7}
\end{table}

\begin{table}[!ht]
\centering
\caption{Spearman's rank correlation coefficient found for each base model's 14 days forecasts.}
\resizebox{0.8\columnwidth}{!}{
{
\begin{tabular}{l|c|c|c|c}
                              & CV & SVDE & KPSS & ACF   \\ 
\bottomrule\toprule
\multicolumn{1}{l|}{ARIMA}   & 0.38  & \multicolumn{1}{c|}{0.27} & 0.51  & \multicolumn{1}{c}{0.34} \\
\multicolumn{1}{l|}{HW}      & 0.29  & \multicolumn{1}{c|}{0.27} & 0.38  & \multicolumn{1}{c}{0.13} \\
\multicolumn{1}{l|}{Prophet} & 0.35  & \multicolumn{1}{c|}{0.25} & 0.45  & \multicolumn{1}{c}{0.12} \\
\multicolumn{1}{l|}{LSTM}    & 0.49  & \multicolumn{1}{c|}{0.31} & 0.49  & \multicolumn{1}{c}{0.11} \\ 
\bottomrule
\end{tabular}
}
}
\label{tab:table_spearman_14}
\end{table}

\subsection{Forecast models}
We implemented all algorithms in Python and a brief summary follows.

\subsubsection{Base models}
The Pyramid ARIMA (pmdarima) API \cite{pmdarima} was used with a grid search configured to estimate the $p$ and $q$ parameters by minimising an Akaike information criterion (AIC) for the nonseasonal curves. The exponential smoothing from statsmodels \cite{seabold2010statsmodels} was implemented with the parameters optimised using a Basin-hopping optimiser. The Prophet method was configured for a linear growth with its changepoints parameter experimentally configured to $14$. A deep LSTM model was implemented using the TensorFlow \cite{abadi2016tensorflow} Keras module with a hidden layer of $192$ units, followed by two $384$-unit layers. We use an Adam optimiser on the LSTM with a low learning rate of $10^{-5}$ due to oscillation found in the losses. The LSTM setup is configured to train over $110$ epochs with a sequence containing batches of temporal data. The batches are computed with the Keras TimeseriesGenerator, configured to use the training data to output a sequence with a length of eight (required number of prior observations). The number of samples for each batch was set to four.

\subsubsection{Stacking models}
Two MLP models were used as second-layer learning algorithms (stacking and feature-weighted stacking) using the TensorFlow Keras module. Their input vectors comprise the two best performing models from the preliminary forecasting stage and the two meta-features with the highest average Spearman's rank correlation coefficient for the feature-weighted stacking model. We used Bayesian optimisation using the Gaussian process method from the scikit-optimize API \cite{2020SciPy-NMeth} with minimisation configured for the meta-learner's sMAPE score to find an initial set of model parameters. Additional hand-tuning allowed a further reduction of a mean absolute error (MAE).

The final MLP architecture is a neural network with two hidden layers, each with $176$ units. This same architecture is used for all of the stacking methods, and they are fit for $199$ epochs at a learning rate of  $3.6 \times 10^{-05}$.

\subsubsection{Model averaging}
Model averaging \cite{claeskens2008model} is an effective ensembling approach that averages the forecasts from more than one learning algorithm. We averaged the predictions from the Prophet and LSTM Neural Network models that performed the best during their first stage of forecasts for the stacking approach. 
The forecasts of the model averaging might thus be expressed by the following simple notation:
\begin{align}
\hat{y}_t = \frac{{y_1}_t + {y_2}_t}{2}
\end{align}

where $y_1$ and $y_2$ are the predictions from the selected forecasting algorithms.

\begin{table}[!htb]
\centering
\caption{sMAPE scores for 7 and 14-day forecast horizons.}\label{tab:results}
\resizebox{0.8\columnwidth}{!}{
{
\begin{tabular}{cccc}
\multicolumn{1}{l|}{}                           & \multicolumn{1}{l|}{7 Days}    & \multicolumn{1}{l}{14 Days} \\ 
\bottomrule\toprule
\multicolumn{1}{l|}{ARIMA}                     & \multicolumn{1}{c|}{4.673}      & \multicolumn{1}{c}{1.109}  \\ 
\multicolumn{1}{l|}{Holt-Winters}              & \multicolumn{1}{c|}{0.792}      & \multicolumn{1}{c}{1.070}  \\ 
\multicolumn{1}{l|}{Prophet}                   & \multicolumn{1}{c|}{0.734}      & \multicolumn{1}{c}{0.898}  \\ 
\multicolumn{1}{l|}{LSTM Neural Network}       & \multicolumn{1}{c|}{0.757}      & \multicolumn{1}{c}{1.906} \\ 
\multicolumn{1}{l|}{Model Averaging}           & \multicolumn{1}{c|}{0.497}      & \multicolumn{1}{c}{1.117} \\ 
\multicolumn{1}{l|}{Stacking}             & \multicolumn{1}{c|}{0.477}      & \multicolumn{1}{c}{1.160} \\ 
\multicolumn{1}{l|}{Feature-weighted stacking} & \multicolumn{1}{c|}{\bf{0.469}} & \multicolumn{1}{c}{\bf{0.816}} \\ 
\bottomrule
\end{tabular}
}
}
\label{tab::scores}
\end{table}

Our results are the average outcomes for the set of out-of-sample countries, and we note that the proposed feature weighted stacking method on average performs best across both time horizons. Although the utility of meta-features produced a slight improvement over the conventional stacking approach for the short-term forecasts, we note that the method scales well as the only successful ensemble for the longer horizon forecasts. 
In contrast, the model averaging and non-feature-weighted stacking methods were outperformed by all of the base models for the longer horizon forecast; hence we find that the proposed method is generally a more robust ensembling approach. The results show that a stacking method benefits from utilising time-series statistics as meta-features since they allow the meta-learner to weigh the model predictions according to their learned performance for differently characterised curves. 

There were deviations in the ensembles' predictions due to the cascading effect of the random initialisations at the different levels of the multi-layer ensembles. Therefore the results were averaged over five runs and recorded in table \ref{tab:results}. Figures \ref{fig::egypt} and \ref{fig::netherlands} illustrate the forecasting results for Egypt and the Netherlands for the two forecasting horizons.

%%%%%%%%%%%%%%%%%%%%%%%%%%%%%%%%%%%%%%%%%%%%%%%%%%%%%%%%%%%%%%%
%                         CONCLUSION
%%%%%%%%%%%%%%%%%%%%%%%%%%%%%%%%%%%%%%%%%%%%%%%%%%%%%%%%%%%%%%%%%

\section{Conclusion}
In this paper, we studied a feature-weighted stacking approach for the forecasting of nonseasonal time-series. We introduced four features that might improve the accuracy of a stacked generalisation when they are utilised as meta-features. The CV and KPSS features showed the highest correlation with the model outcomes. They were both included in the feature-weighted stacking ensembles. These ensembles consisted of the LSTM Neural Network and Prophet models for both forecast horizons.

The proposed method produced an average sMAPE score of 0.469 and 0.816 for the seven and fourteen-day forecasts. Although these forecasts only beat the scores of others by a small margin, the results are conclusive that the methodology is a more favourable alternative than model averaging and conventional stacking. 

It is common in forecasting to adopt simple model averaging to combine the strengths of multiple forecasting models. However, our results suggest that although model averaging remains an effective technique, we can produce improved forecasts using the proposed feature-weighted stacking method. Moreover, we have shown that using meta-features is relevant to the forecasting problem. 

We only presented results for using neural networks as a meta-learner, and future work should assess more algorithms. Further research is also required to evaluate additional features that may, under certain circumstances, add value when included in ensembles. Further work could also determine the method’s performance against that of model selection and other ensembling strategies.

\bibliographystyle{IEEEtran}
\bibliography{bib}

\begin{thebibliography}{10}
\providecommand{\url}[1]{#1}
\csname url@samestyle\endcsname
\providecommand{\newblock}{\relax}
\providecommand{\bibinfo}[2]{#2}
\providecommand{\BIBentrySTDinterwordspacing}{\spaceskip=0pt\relax}
\providecommand{\BIBentryALTinterwordstretchfactor}{4}
\providecommand{\BIBentryALTinterwordspacing}{\spaceskip=\fontdimen2\font plus
\BIBentryALTinterwordstretchfactor\fontdimen3\font minus
  \fontdimen4\font\relax}
\providecommand{\BIBforeignlanguage}[2]{{%
\expandafter\ifx\csname l@#1\endcsname\relax
\typeout{** WARNING: IEEEtran.bst: No hyphenation pattern has been}%
\typeout{** loaded for the language `#1'. Using the pattern for}%
\typeout{** the default language instead.}%
\else
\language=\csname l@#1\endcsname
\fi
#2}}
\providecommand{\BIBdecl}{\relax}
\BIBdecl

\bibitem{van2021did}
T.~L. van Zyl and T.~Celik, ``Did we produce more waste during the covid-19
  lockdowns? a remote sensing approach to landfill change analysis,''
  \emph{IEEE Journal of Selected Topics in Applied Earth Observations and
  Remote Sensing}, vol.~14, pp. 7349--7358, 2021.

\bibitem{perumal2020comparison}
R.~Perumal and T.~L. van Zyl, ``Comparison of recurrent neural network
  architectures for wildfire spread modelling,'' in \emph{2020 International
  SAUPEC/RobMech/PRASA Conference}.\hskip 1em plus 0.5em minus 0.4em\relax
  IEEE, 2020, pp. 1--6.

\bibitem{wolpert1997no}
D.~H. Wolpert and W.~G. Macready, ``No free lunch theorems for optimization,''
  \emph{IEEE transactions on evolutionary computation}, vol.~1, no.~1, pp.
  67--82, 1997.

\bibitem{ribeiro2020short}
M.~H. D.~M. Ribeiro, R.~G. da~Silva, V.~C. Mariani, and L.~dos Santos~Coelho,
  ``Short-term forecasting covid-19 cumulative confirmed cases: Perspectives
  for brazil,'' \emph{Chaos, Solitons \& Fractals}, p. 109853, 2020.

\bibitem{atherfold2020method}
J.~Atherfold and T.~L. Van~Zyl, ``A method for dissolved gas forecasting in
  power transformers using ls-svm,'' in \emph{2020 IEEE 23rd International
  Conference on Information Fusion (FUSION)}.\hskip 1em plus 0.5em minus
  0.4em\relax IEEE, 2020, pp. 1--8.

\bibitem{MakridakisComparisons}
S.~Makridakis, E.~Spiliotis, and V.~Assimakopoulos, ``Statistical and machine
  learning forecasting methods: Concerns and ways forward,'' \emph{PLoS ONE},
  vol.~13, 03 2018.

\bibitem{ribeiro2020ensemble}
M.~H. D.~M. Ribeiro and L.~dos Santos~Coelho, ``Ensemble approach based on
  bagging, boosting and stacking for short-term prediction in agribusiness time
  series,'' \emph{Applied Soft Computing}, vol.~86, p. 105837, 2020.

\bibitem{makridakis2020m4}
S.~Makridakis, E.~Spiliotis, and V.~Assimakopoulos, ``The m4 competition:
  100,000 time series and 61 forecasting methods,'' \emph{International Journal
  of Forecasting}, vol.~36, no.~1, pp. 54--74, 2020.

\bibitem{coscrato2020nn}
V.~Coscrato, M.~H. de~Almeida~In{\'a}cio, and R.~Izbicki, ``The nn-stacking:
  Feature weighted linear stacking through neural networks,''
  \emph{Neurocomputing}, 2020.

\bibitem{montero2020fforma}
P.~Montero-Manso, G.~Athanasopoulos, R.~J. Hyndman, and T.~S. Talagala,
  ``Fforma: Feature-based forecast model averaging,'' \emph{International
  Journal of Forecasting}, vol.~36, no.~1, pp. 86--92, 2020.

\bibitem{dong2020interactive}
E.~Dong, H.~Du, and L.~Gardner, ``An interactive web-based dashboard to track
  covid-19 in real time,'' \emph{The Lancet infectious diseases}, vol.~20,
  no.~5, pp. 533--534, 2020.

\bibitem{barak2019time}
S.~Barak, M.~Nasiri, and M.~Rostamzadeh, ``Time series model selection with a
  meta-learning approach; evidence from a pool of forecasting algorithms,''
  \emph{arXiv preprint arXiv:1908.08489}, 2019.

\bibitem{caraiani2014predictive}
P.~Caraiani, ``The predictive power of singular value decomposition entropy for
  stock market dynamics,'' \emph{Physica A: Statistical Mechanics and its
  Applications}, vol. 393, pp. 571--578, 2014.

\bibitem{kwiatkowski1992testing}
D.~Kwiatkowski, P.~C. Phillips, P.~Schmidt, and Y.~Shin, ``Testing the null
  hypothesis of stationarity against the alternative of a unit root: How sure
  are we that economic time series have a unit root?'' \emph{Journal of
  econometrics}, vol.~54, no. 1-3, pp. 159--178, 1992.

\bibitem{fattah2018forecasting}
J.~Fattah, L.~Ezzine, Z.~Aman, H.~El~Moussami, and A.~Lachhab, ``Forecasting of
  demand using arima model,'' \emph{International Journal of Engineering
  Business Management}, vol.~10, p. 1847979018808673, 2018.

\bibitem{kalekar2004time}
P.~S. Kalekar \emph{et~al.}, ``Time series forecasting using holt-winters
  exponential smoothing,'' \emph{Kanwal Rekhi School of Information
  Technology}, vol. 4329008, no.~13, pp. 1--13, 2004.

\bibitem{taylor2018forecasting}
S.~J. Taylor and B.~Letham, ``Forecasting at scale,'' \emph{The American
  Statistician}, vol.~72, no.~1, pp. 37--45, 2018.

\bibitem{mathonsi2020prediction}
T.~Mathonsi and T.~L. van Zyl, ``Prediction interval construction for
  multivariate point forecasts using deep learning,'' in \emph{2020 7th
  International Conference on Soft Computing \& Machine Intelligence
  (ISCMI)}.\hskip 1em plus 0.5em minus 0.4em\relax IEEE, 2020, pp. 88--95.

\bibitem{sak2014long}
H.~Sak, A.~Senior, and F.~Beaufays, ``Long short-term memory based recurrent
  neural network architectures for large vocabulary speech recognition,''
  \emph{arXiv preprint arXiv:1402.1128}, 2014.

\bibitem{agirre2006regression}
E.~Agirre-Basurko, G.~Ibarra-Berastegi, and I.~Madariaga, ``Regression and
  multilayer perceptron-based models to forecast hourly o3 and no2 levels in
  the bilbao area,'' \emph{Environmental Modelling \& Software}, vol.~21,
  no.~4, pp. 430--446, 2006.

\bibitem{pmdarima}
\BIBentryALTinterwordspacing
T.~G. Smith \emph{et~al.}, ``{pmdarima}: Arima estimators for {Python},''
  2017--. [Online]. Available: \url{http://www.alkaline-ml.com/pmdarima}
\BIBentrySTDinterwordspacing

\bibitem{seabold2010statsmodels}
S.~Seabold and J.~Perktold, ``statsmodels: Econometric and statistical modeling
  with python,'' in \emph{9th Python in Science Conference}, 2010.

\bibitem{abadi2016tensorflow}
M.~Abadi, P.~Barham, J.~Chen, Z.~Chen, A.~Davis, J.~Dean, M.~Devin,
  S.~Ghemawat, G.~Irving, M.~Isard \emph{et~al.}, ``Tensorflow: A system for
  large-scale machine learning,'' in \emph{12th $\{$USENIX$\}$ symposium on
  operating systems design and implementation ($\{$OSDI$\}$ 16)}, 2016, pp.
  265--283.

\bibitem{2020SciPy-NMeth}
P.~{Virtanen}, R.~{Gommers}, T.~E. {Oliphant}, M.~{Haberland}, T.~{Reddy},
  D.~{Cournapeau}, E.~{Burovski}, P.~{Peterson}, W.~{Weckesser}, J.~{Bright},
  S.~J. {van der Walt}, M.~{Brett}, J.~{Wilson}, K.~{Jarrod Millman},
  N.~{Mayorov}, A.~R.~J. {Nelson}, E.~{Jones}, R.~{Kern}, E.~{Larson},
  C.~{Carey}, {\.I}.~{Polat}, Y.~{Feng}, E.~W. {Moore}, J.~{Vand erPlas},
  D.~{Laxalde}, J.~{Perktold}, R.~{Cimrman}, I.~{Henriksen}, E.~A. {Quintero},
  C.~R. {Harris}, A.~M. {Archibald}, A.~H. {Ribeiro}, F.~{Pedregosa}, P.~{van
  Mulbregt}, and S.~.~. {Contributors}, ``{SciPy 1.0: Fundamental Algorithms
  for Scientific Computing in Python},'' \emph{Nature Methods}, vol.~17, pp.
  261--272, 2020.

\bibitem{claeskens2008model}
G.~Claeskens, N.~L. Hjort \emph{et~al.}, ``Model selection and model
  averaging,'' \emph{Cambridge Books}, 2008.

\end{thebibliography}

\end{document}